\documentclass[a4paper, 10pt, conference]{ieeeconf}      

\IEEEoverridecommandlockouts                              

\usepackage{graphics} 
\usepackage{epsfig} 
\usepackage{mathptmx} 
\usepackage{times} 
\usepackage{amsmath} 
\usepackage{amssymb}  
\usepackage{subcaption}
\usepackage{bm}
\usepackage{multirow}

\title{\LARGE \bf
Safe Speed Control and Collision Probability Estimation Under Ego-Pose Uncertainty for Autonomous Vehicle}

\author{Vladislav Kibalov$^{1}$ and Oleg Shipitko$^{2}$
\thanks{$^{1}$ Institute for Information Transmission Problems – IITP RAS, Bol’shoy Karetnyy Pereulok 19, Moscow, Russia, 127051; {\tt\small shipitko@iitp.ru}}%
}

\begin{document}

\maketitle
\thispagestyle{empty}
\pagestyle{empty}

\begin{abstract}
In order for autonomous vehicles to become a part of the Intelligent Transportation Ecosystem, they are required to guarantee a particular level of safety. For that to happen a safe vehicle control algorithms need to be developed, which include assessing the probability of a collision while driving along a given trajectory and selecting control signals that minimize this probability.
In this paper, we propose a speed control system that estimates a collision probability taking into account static and dynamic obstacles as well as ego-pose uncertainty and chooses the maximum safe speed. For that, the planned trajectory is converted by the control system into control signals that form input for the dynamic vehicle model. The model predicts a real vehicle path. The predicted trajectory is generated for each particle -- a weighted by a probability hypothesis of the localization system about the vehicle pose. Based on the predicted particles' trajectories, the probability of collision is calculated, and a decision is made on the maximum safe speed. The proposed algorithm was validated on the real autonomous vehicle. The experimental results demonstrate that the proposed speed control system reduces the vehicle speed to a safe value when performing maneuvers and driving through narrow openings. Therefore the observed behavior of the system is mimicking a human driver behavior when driving in difficult and ambiguous traffic situations.
\end{abstract}

\section{INTRODUCTION}

The active development of the autonomous vehicles brings closer their integration into the urban transportation system. One of the main advantages of autonomous vehicles over traditional human-driven cars is an increased driving safety~\cite{harper2016cost, liu2018safe, wang2019crash} or, in other words, a decrease in the number of traffic accidents. However, in reality, there are many technical and scientific problems, to overcome to guarantee the safety of autonomous vehicles. One of these problems is the safe vehicle control, which comprises assessing the collision probability while driving along a given trajectory and selecting control signals that minimize this probability~\cite{lambert2008fast}.

Many existing approaches to motion planning and the collision probability estimation assume the precisely determined vehicle ego-pose~\cite{hu2018dynamic, kuffner2000rrt}. However, when working in a real environment, it is crucial to take into account the pose uncertainty caused by unpredictable deviations in vehicle motion, inaccuracies in sensor measurements, changing environment, and other uncertainty factors. Thus, the task of assessing the safety of the motion trajectory under conditions of ego-pose uncertainty is of high priority, which is confirmed by the great number of research dedicated to this problem (e.g.~\cite{bopardikar2015multiobjective, du2011probabilistic}).

\begin{figure}[thpb]
    \centering
    \includegraphics[width=0.8\linewidth]{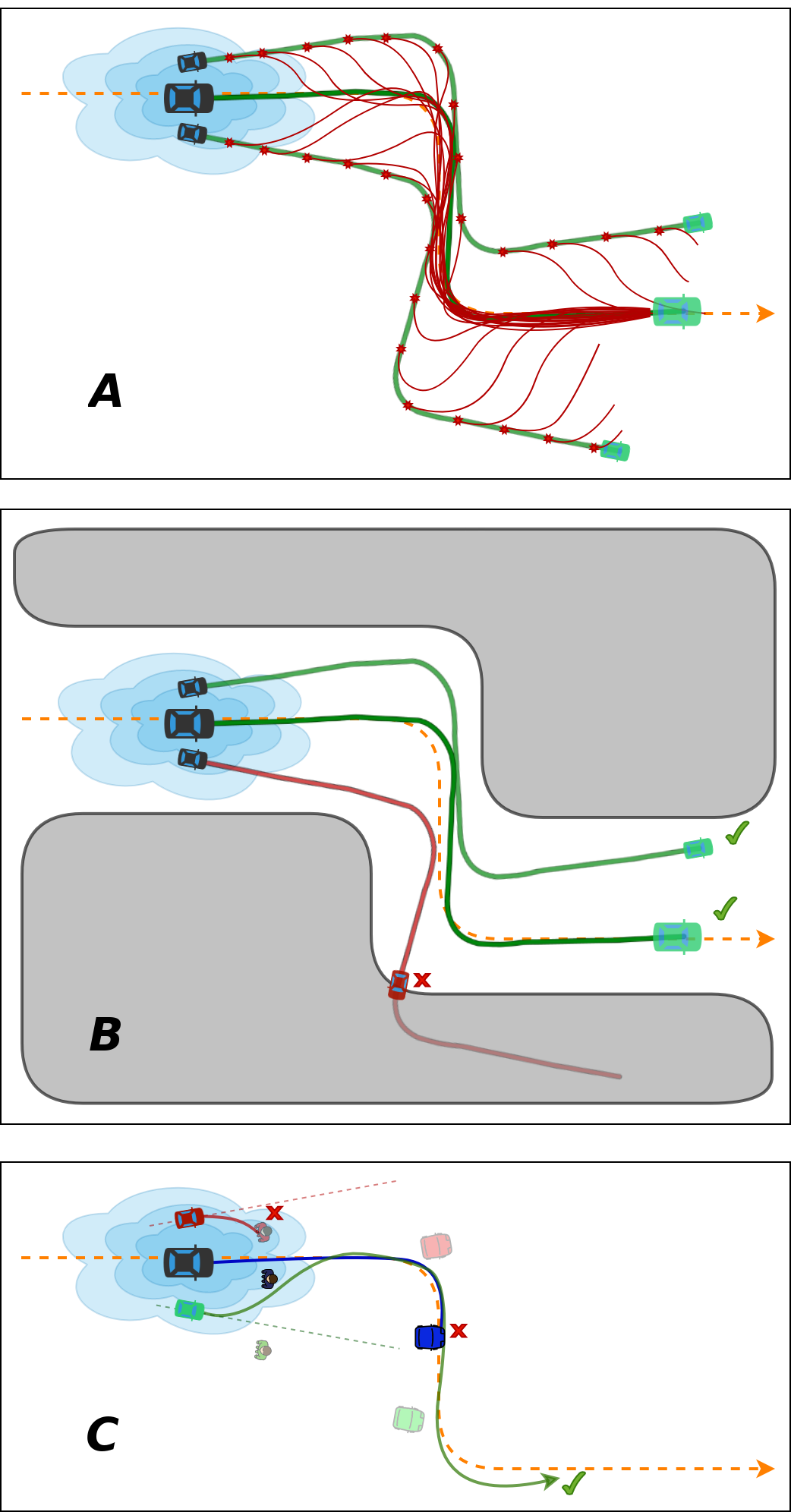}
    \caption{Approximation of the space distribution of the future trajectories (\textbf{A}), method to avoid a collision with obstacles from static occupancy grid map (\textbf{B}), method to avoid a collision with dynamic obstacles (\textbf{C}).}
    \label{fig:key_consideration}
\end{figure}

A common approach to motion planning and collision probability estimation is Monte Carlo Motion Planning. It estimates collision probability by repeatedly simulating the vehicle movement along the desired trajectory~\cite{janson2018monte, schmerling2016evaluating, broadhurst2005monte}. The probability is calculated as the ratio of the number of simulations in which the fact of a collision was recorded to the total number of simulations. This approach requires a large number of experiments to obtain a reliable estimate of the probability, and, as a result, is computationally complex.

An alternative approach is based on the prior vehicle state-space probability distribution propagation along the predefined trajectory~\cite{liu2014incremental, van2011lqg, vitus2011closed, houenou2014risk}.
In~\cite{patil2012estimating}, the authors propose an extension to this approach, taking into account that the distribution of states at each moment depends on whether the distributions at previous time moments were free from collisions. Accounting is carried out through truncating the distribution at each step of the algorithm -- discarding hypotheses for which a collision is observed, and approximating the truncated distribution by Gaussian one. Such an approach makes it possible to propagate the truncated probability distribution forward along the trajectory and, as a result, makes it possible to calculate the collision probability more precisely. The disadvantage of all algorithms based on the propagation of prior distribution is the approximation of sensors measurement errors and the uncertainty of vehicle motion by the normal distribution. Such an approximation allows one to calculate the collision probability analytically, however, it is rarely observed in real-world problems.

Some works consider the problem of trajectory safety assessment, not only taking into account static obstacles but also considering possible trajectories of other road users~\cite{schreier2016integrated, mehta2018c, annell2016probabilistic, lienke2018ad}.

Even though the tasks of trajectory planning and the collision probability estimation are widely studied, most existing approaches do not take into account the uncertainty of vehicle ego-pose at the time of planning. The vehicle’s current position is considered precisely known, and future uncertainty is usually modeled by a normal distribution centered in the points of the desired trajectory, which does not reflect a real motion uncertainty.

A vehicle speed control system is proposed in this work. It controls the speed to achieve an acceptable value of the collision probability which is imposed by the probability threshold function.
To assess the safety of the vehicle trajectory under conditions of pose uncertainty the proposed system accepts as an input an arbitrary probability distribution of vehicle ego-pose, which also distinguishes it from existing approaches. Unlike other approaches Monte Carlo method is not used directly to assess the safety of the trajectory, but only used to estimate the current probability distribution of the vehicle position for the evaluation of proposed system. Although the Monte Carlo localization method is used as an ego-pose probability source the system can work with any other type of ego-pose probability distribution (e.g. Kalman filter).
To assess safety, only a part of the future trajectory, predicted by a vehicle dynamic model, is used. This significantly reduces computational load and allows the proposed approach to be used in real-time systems.

There are two main contributions presented in this paper. The first contribution is the novel safe speed control system which explicitly accounts for the vehicle ego-pose uncertainty. The second one is the method for collision probability estimation. The proposed method distinguishes uncertainty handling for static obstacles specified in the global reference frame and for dynamic obstacles which are detected in the vehicle local reference frame.



\section{PROBLEM STATEMENT}
In this paper, we propose the Safe Speed Control System Under Ego-Pose Uncertainty for Autonomous Vehicle. The system has the following inputs:
\begin{itemize}

\item estimated vehicle pose $\mathbf{X}^t$ in map reference frame at the current moment of time $t$;

\item probability distribution of vehicle pose \\ $\mathbf{M} = \{ (\mathbf{X}^t_i , w^t_i) \}$ , where $w^t_i$ is the probability of $\mathbf{X}^t_i$ to be an ego-pose; 

\item current vehicle speed $V^t$;

\item reference trajectory $\mathbf{T}$ -- the sequence of waypoints i.e. global trajectory;

\item static occupancy grid map $\mathbf{M_{oc}}$ as a binary image;

\item a set of detected dynamic obstacles $\mathbf{D}$, where each obstacle is a polyline that circumscribes the projection of an obstacle to the road plane in the reference frame attached to the vehicle. 
\end{itemize}

The output of the system is the maximum safe speed $V_s$, such that:
\begin{equation}\label{eq:safe_speed_lim}
\begin{array}{l}
    V_s = \mathrm{max} \big( V_{\mathrm{lim}} \in [0 ; V_{\mathrm{max}}] | P(C | \mathbf{X}^t, \mathbf{M}, V^t, \mathbf{T}, \mathbf{M_{oc}}, \\ 
                                                   \mathbf{D}, \tau, V_{\mathrm{lim}}) < P_s(V_{\mathrm{lim}}) \big),
\end{array}
\end{equation}
where $V_{\mathrm{max}}$ is the maximum possible speed (restricted by road properties), $P_s(V_{\mathrm{lim}})$ -- collision probability threshold function, $P(C|\ldots)$ (further referred as to $P_C$) -- conditional vehicle collision probability for given $V_{\mathrm{lim}}$ within the prediction horizon [$t$, $t+\tau$], $\tau$ -- prediction duration, $V_{\mathrm{lim}}$ is a speed limit for prediction horizon -- the maximum allowed speed on the predicted trajectory. 

Simply put, the objective of equation~(\ref{eq:safe_speed_lim}) is to determine $V_s$ which is the highest speed limit among considered speed limits $V_{\mathrm{lim}}$ that is considered to be safe, i.e. conditional collision probability $P_C$ is less than collision probability threshold determined by $P_s(V_{\mathrm{lim}})$ function within prediction horizon.

\section{KEY ASSUMPTIONS}

In order to estimate collision probability within prediction horizon, it is necessary to analyze the space of all possible vehicle future trajectories.
The variability of future trajectories firstly depends on ego-pose uncertainty determined in the map reference frame at the current moment of time, and secondly on how this uncertainty will change over time.
Possible future changes of estimated pose will lead to variations of future trajectory shape due to the shift of new pose estimation relative to the reference trajectory. Such changes are impossible to predict. We assume that they may occur randomly at a random moment, then the best approximation of the spatial distribution of the future trajectories is shown in Fig.~\ref{fig:key_consideration} (\textbf{A}). In this figure, green lines are the copies of the predicted trajectory for current estimated vehicle pose $\mathbf{X}^t$. Each of them is transferred to pose of every hypothesis $\mathbf{X}^t_i$, in other words, they capture the future motion of pose distribution in case of absence of new sensory data. Red lines are possible trajectories provided that a random hypothesis became the new estimated pose at a random moment of time.

Since it is computationally intractable to generate so many predictions of trajectories with sufficient discretization, several simplifications are proposed.

In order to avoid a collision with static obstacles we can predict trajectory for estimated vehicle pose and transfer it to pose of each hypothesis without varying its shape (same as green lines from Fig.~\ref{fig:key_consideration} \textbf{A}). This approach allows us to assess collision probability within prediction horizon in case of a complete lack of new data to reestimate pose distribution, which represents the worst-case scenario. If the collision probability for the worst-case scenario does not exceed the threshold value, then it will not exceed the threshold in any other case. Schematically, this approach is illustrated in Fig.~\ref{fig:key_consideration} (\textbf{B}).

With dynamic obstacles, this approach is not suitable, since their coordinates are measured relative to the vehicle and for each hypothesis, the relative position of the obstacles will be the same (provided that the error of the obstacle detector is neglected). To estimate the probability, we can consider the case in which the variation in the relative profiles of predicted trajectories will be maximum. This case arises if the trajectories will be predicted for each hypothesis separately, taking it as the estimated vehicle pose. Schematically, this approach is illustrated in Fig.~\ref{fig:key_consideration} (\textbf{C}).

\section{SAFE SPEED CONTROL SYSTEM STRUCTURE}

\begin{figure}[thpb]
    \centering
    \includegraphics[width=\linewidth]{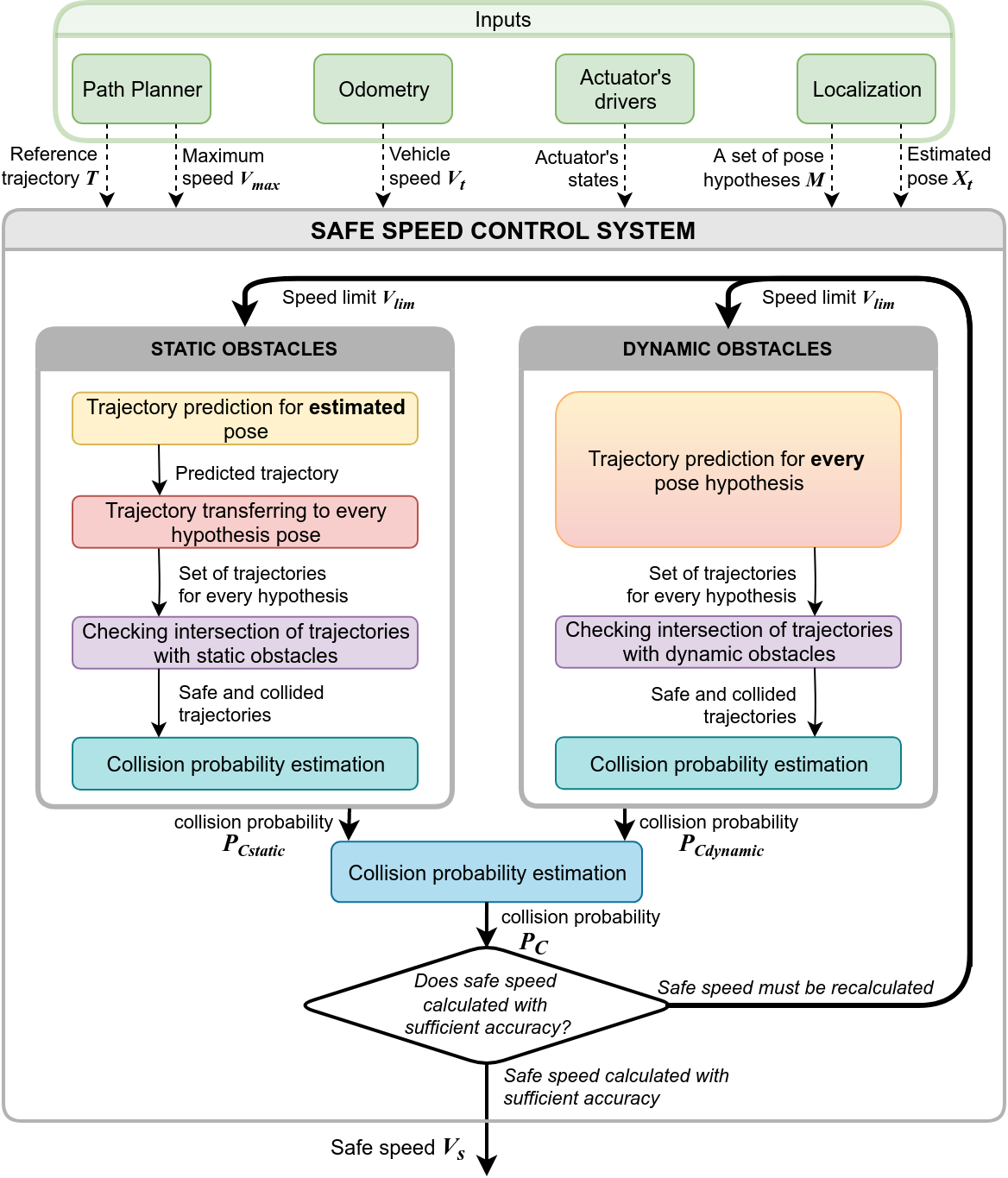}
    \caption{Vehicle Safe Speed Control System flowchart.}
    \label{fig:common_scheme}
\end{figure}

The system structure is shown on Fig.~\ref{fig:common_scheme}. 
Safe speed $V_s$ is recalculated periodically. The calculation is performed in cycle, where at each cycle iteration collision probability is estimated for given  $V_{lim}$. The collision probability is determined by two distinct probabilities: the probability of collision with static $ P_{\mathrm{Cstatic}} $ and with dynamic $ P_{\mathrm{Cdynamic}} $ obstacles (both described in the respective sections).

On a lower level, main operations inside the Safe Speed Control System are:

\begin{itemize}

\item Trajectory prediction, described in \textbf{Trajectory Prediction Module} section. 

\item Checking intersection of trajectories with static obstacles $\mathbf{M_{oc}}$.

\item Checking intersection of trajectories with dynamic obstacles $\mathbf{D}$.

\item Collision probability estimation based on information about collision for every trajectory. For static obstacles it described in section \textbf{Collision Probability Estimation With Static Obstacles}. With dynamic obstacles, it works in a similar way.

\item Safe speed estimation, i.e. determination of optimal speed limit $V_s$ based on the calculated collision probabilities $ P_C $ for various speed limits $ V_{\mathrm{lim}} $ (according to the equation~(\ref{eq:safe_speed_lim})). Described in details in \textbf{Safe Speed Estimation} section.

\end{itemize}

\section{TRAJECTORY PREDICTION MODULE}
\begin{figure}[thpb]
\centering
    \includegraphics[width=\linewidth]{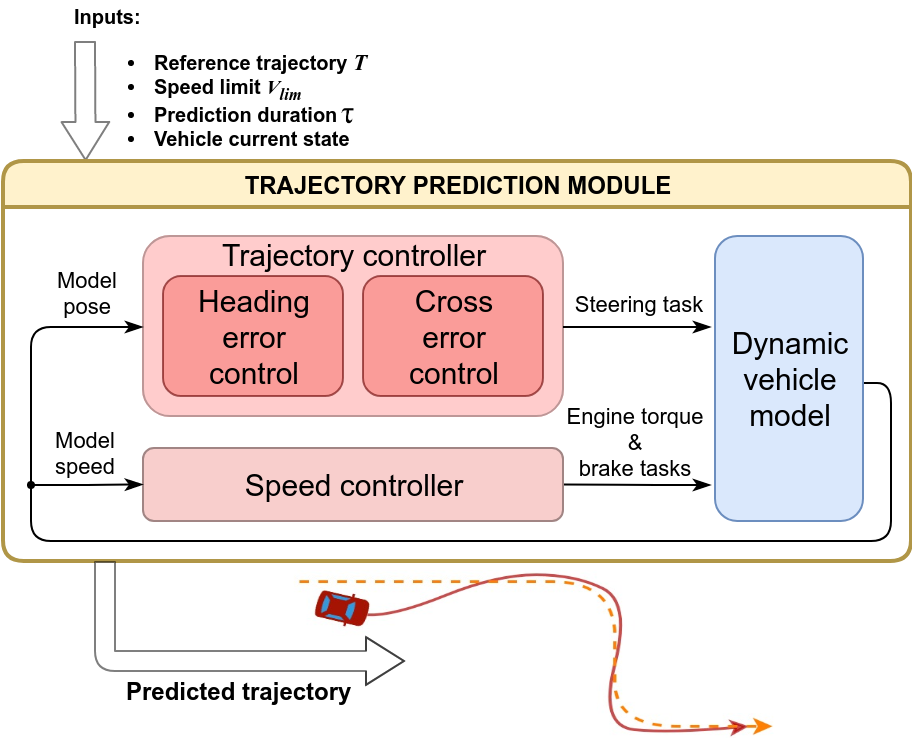}
    \caption{Trajectory Prediction Module flowchart.}
    \label{fig:predictor_scheme}
\end{figure}

Trajectory Prediction Module calculates the future trajectory in real-time based on the dynamic model of the vehicle. The inputs are vehicle current state (pose $\mathbf{X}^t$, speed $V^t$ and etc.), reference trajectory $\mathbf{T}$, speed limit $V_{\mathrm{lim}}$, prediction duration $\tau$. The output is predicted trajectory. The module structure is shown in Fig. \ref{fig:predictor_scheme}.

Trajectory Prediction Module combines the vehicle motion control system (CS) consisting of trajectory and speed controllers and a dynamic vehicle model. CS used in this module is fully identical to the one that controls the vehicle. This guarantees the complete identity of the control signals generated by both control systems.

In case of collision probability calculation with static obstacles, a trajectory prediction is calculated for estimated vehicle pose $\mathbf{X}^t$ only. In case of dynamic obstacles, future trajectories are generated for each hypothesis separately, taking its pose $\mathbf{X}^t_i$ as the estimated vehicle pose.

\section{COLLISION PROBABILITY ESTIMATION WITH STATIC OBSTACLES}
The collision probability with static obstacles $ P_{\mathrm{Cstatic}} $ is estimated as follows:
\begin{itemize}

\item Trajectory Prediction Module calculates the future trajectory for the estimated vehicle pose $\mathbf{X}^t$ (\textbf{B} in Fig.~\ref{fig:result_sec}).

\item The trajectory is transferred to each hypothesis so that its beginning coincides with the current hypothesis position, and rotates by the relative yaw angle. Thus, trajectory of each hypothesis is obtained (\textbf{C} in Fig.~\ref{fig:result_sec}).

\item Every trajectory $\mathbf{T}_i$ is checked for collisions with static occupancy grid map $\mathbf{M_{oc}}$. A trajectory is considered safe if at every its point vehicle does not intersect with static obstacles (\textbf{D} in Fig.~\ref{fig:result_sec}).

\item The probability of collision with static obstacles $ P_{\mathrm{Cstatic}} $ is calculated:
\end{itemize}
\begin{equation}
\label{eq:probability}
\begin{array}{l}
    P_{\mathrm{Cstatic}} = \int_\mathbf{M} P(C_{\mathrm{static}}|\mathbf{X} = \mathbf{X}_i) P(\mathbf{X}_i)d\mathbf{X}_i \approx \\
    \approx \frac{\sum_{i=1}^n P(C_{\mathrm{static}}|\mathbf{X} = \mathbf{X}_i) w_i}{\sum_{i=1}^n w_i} =\\
    = \sum_{i=1}^n P(C_{\mathrm{static}}|\mathbf{X} = \mathbf{X}_i) w_i,
\end{array}
\end{equation}
where $ \int_\mathbf{M} d\mathbf{X}_i $ is an integral over pose distribution. The Monte Carlo method approximates this space with a discrete set of particles, so the integral is replaced by the sum over the set of particles; $ n $ is the number of trajectories, each corresponds to the $ i $-th particle; $P(\mathbf{X}_i) = w_i $ is the probability of $ i $-th particle pose to be the current vehicle pose; $ P(C_{\mathrm{static}}|\mathbf{X} = \mathbf{X}_i) $ is the probability of a collision between vehicle projection $\mathbf{R}( \mathbf{X})$ and static obstacles along the trajectory, given that $i$-th particle pose is the current vehicle pose: 
\begin{equation}
P(C_{\mathrm{static}}|\mathbf{X} = \mathbf{X}_i) = \left\{ \begin{array}{ll}
    \text{0,} & \textrm{if } \forall \mathbf{X'}\in    \mathbf{T}_i : \mathbf{R}( \mathbf{X})\cap \mathbf{M_{oc}} =\emptyset \textrm{,}\\
    \text{1,} & \textrm{otherwise}.
 \end{array} \right.
\label{eq:w_om} 
\end{equation}

\section{COLLISION PROBABILITY ESTIMATION WITH DYNAMIC OBSTACLES}
The collision probability with dynamic obstacles $ P_{\mathrm{Cdynamic}} $ is calculated in a similar way, except that for each hypothesis pose $ \mathbf{X}^t_i $ the trajectory prediction module predicts trajectory separately (assuming $ \mathbf{X}^t_i $ to be the vehicle pose). After that, it is the relative profile of each trajectory in the vehicle reference frame that is checked for the presence of collisions with obstacles $ \mathbf{D} $ (also defined in the vehicle reference frame).

\section{SAFE SPEED ESTIMATION}

Total collision probability $ P_C $ with both dynamic and static obstacles can be calculated as follows:
\begin{equation}
P_C = 1 - (1 - P_{\mathrm{Cstatic}}) * (1 - P_{\mathrm{Cdynamic}})
\label{eq:pc_final} 
\end{equation}

\begin{figure}[thpb]
\centering
    \includegraphics[width=\linewidth]{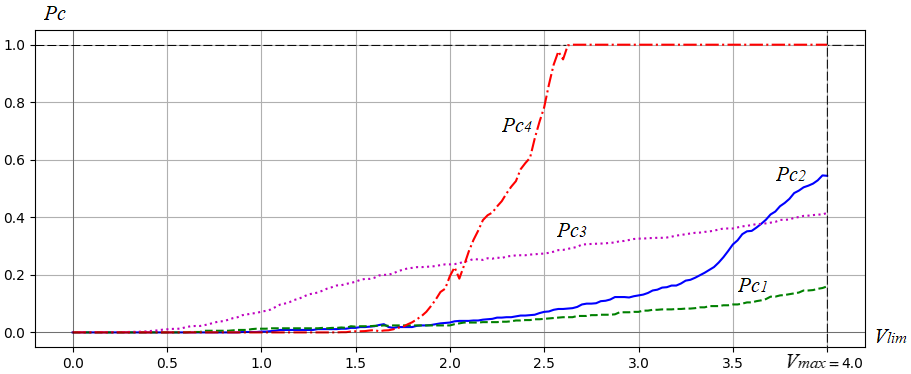}
    \caption{Dependence of collision probability on speed limit}
    \label{fig:collision_probability_chart}
\end{figure}

The resulting probability $ P_C $ is used to calculate the maximum safe speed $ V_s $. The dependence of $ P_C $ on speed limit $ V_{\mathrm{lim}} $ for several arbitrary test cases is shown in Fig.~\ref{fig:collision_probability_chart}. We assume that function $ P_C(V_{\mathrm{lim}}) $ increases monotonically. According to experimental results, it holds in most cases however slight deviations are possible. If we neglect these deviations, then instead of a brute-force search, the calculation of the safe speed $ V_s $ can be performed by quick search algorithms, e.g. binary search, i.e. at each iteration of the algorithm, the obtained collision probability $ P_C $ for the current speed limit $ V_{\mathrm{lim}} $ is compared with the value of the probability threshold function $ P_s $ for the same $ V_{\mathrm{lim}} $. After this, the collision probability $ P_s $ is recalculated for the new speed limit value $ V_{\mathrm{lim}} $ set according to the binary search algorithm from the range $[0 ; V_{\mathrm{max}}]$.

\section{EXPERIMENTAL RESULTS}

The experiments were performed for collision avoidance with static obstacles. The car-like robot was used as an experimental vehicle.

\begin{figure}[thpb]
\centering
    \includegraphics[width=1.0\linewidth]{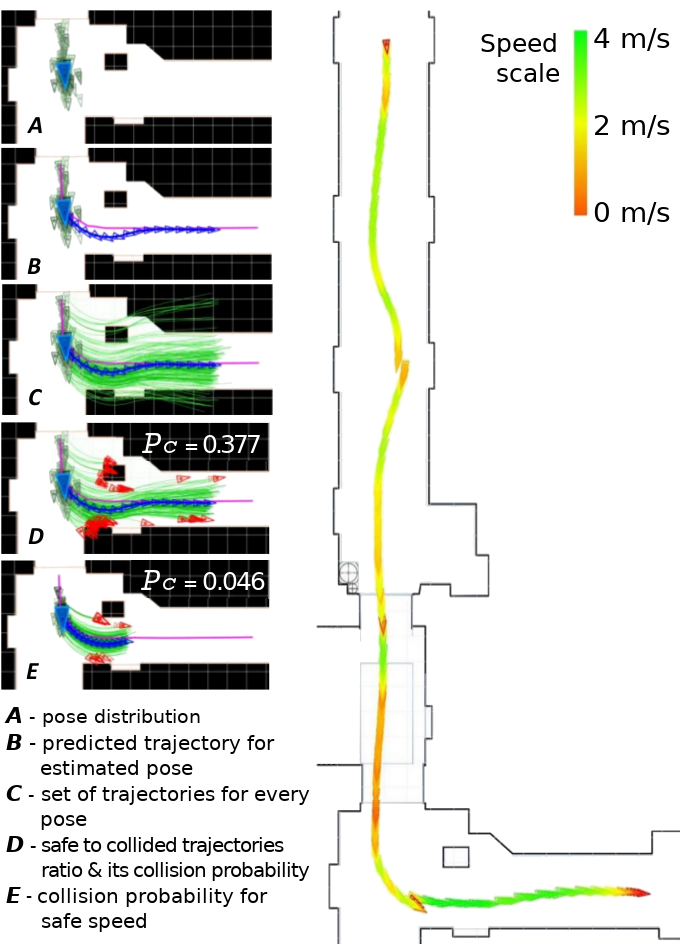}
    \caption{Step-by-step visualization of the safe speed control system operation (left). Vehicle trajectory with speed profile obtained by the speed control system (right).}
    \label{fig:result_sec}
\end{figure}

In Fig.~\ref{fig:result_sec} on the left, the system step-by-step operation is visualized. Here, blue indicates the estimated vehicle pose and predicted trajectory, dark green indicates particles -- pose hypotheses, light green lines -- future trajectories for each particle. Trajectories ending with a red triangle have collisions.

In Fig.~\ref{fig:result_sec} on the right, the vehicle trajectory obtained as a result of real passage is shown. The trajectory color at each point corresponds to the instantaneous speed of the vehicle. Red color corresponds to zero speed, green - to the maximum possible speed $V_{\mathrm{max}}$. The discontinuities observed in the trajectory arise due to corrections of estimated vehicle pose by the localization system. It can be seen from the figure that the speed was severely limited in places where the pose estimation was not accurate, the movement occurred close to the walls or in front of difficult sections of the route such as turns or narrow entrances. After clarifying vehicle estimated pose or after passing difficult sections, the speed was recovered.
 
\begin{figure}[thpb]
\centering
    \includegraphics[width=\linewidth]{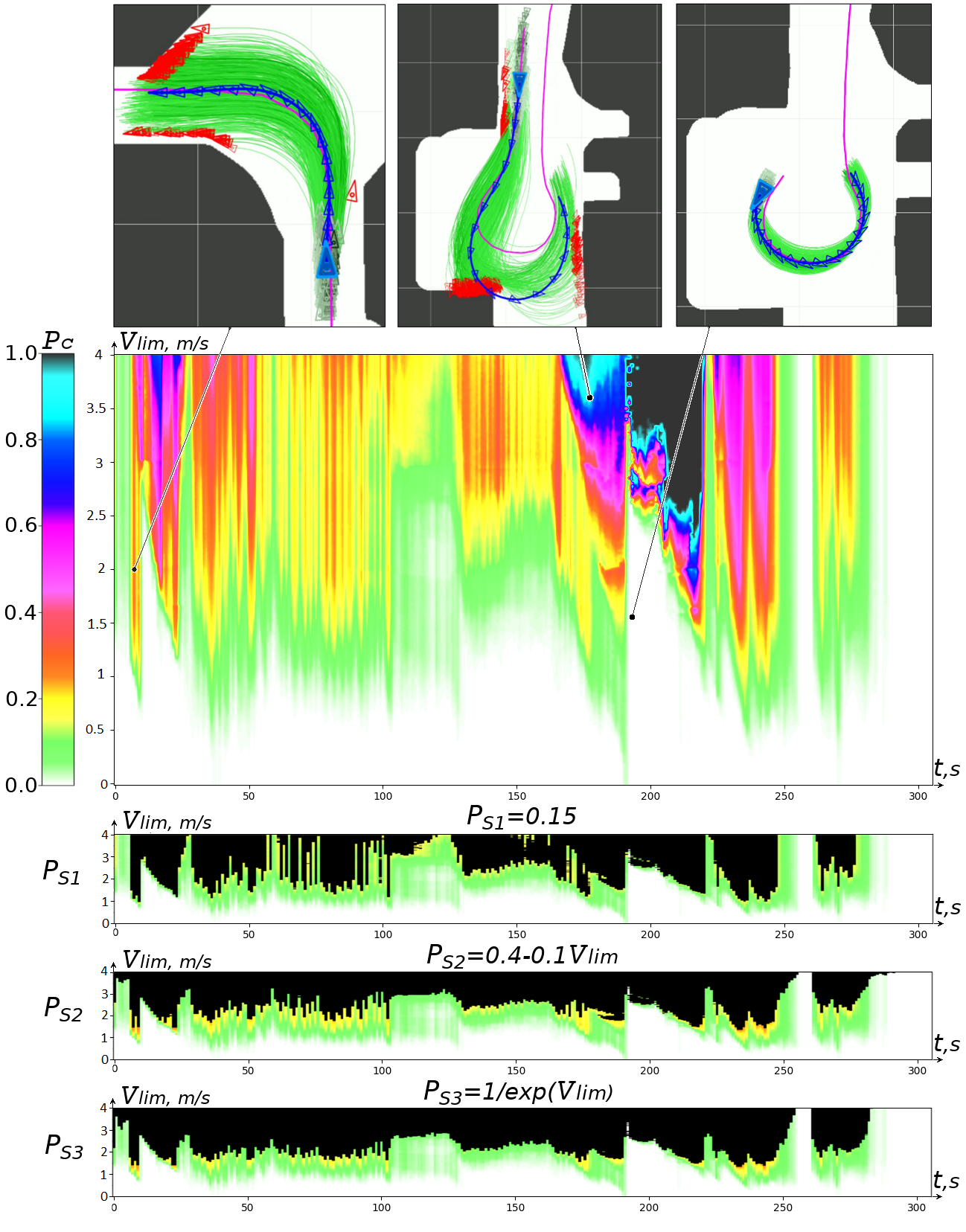}
    \caption{Dependence of collision probability on speed limit over time.}
    \label{fig:result_main}
\end{figure}

\begin{figure*}[thpb]
\centering
    \includegraphics[width=1.0\textwidth,keepaspectratio]{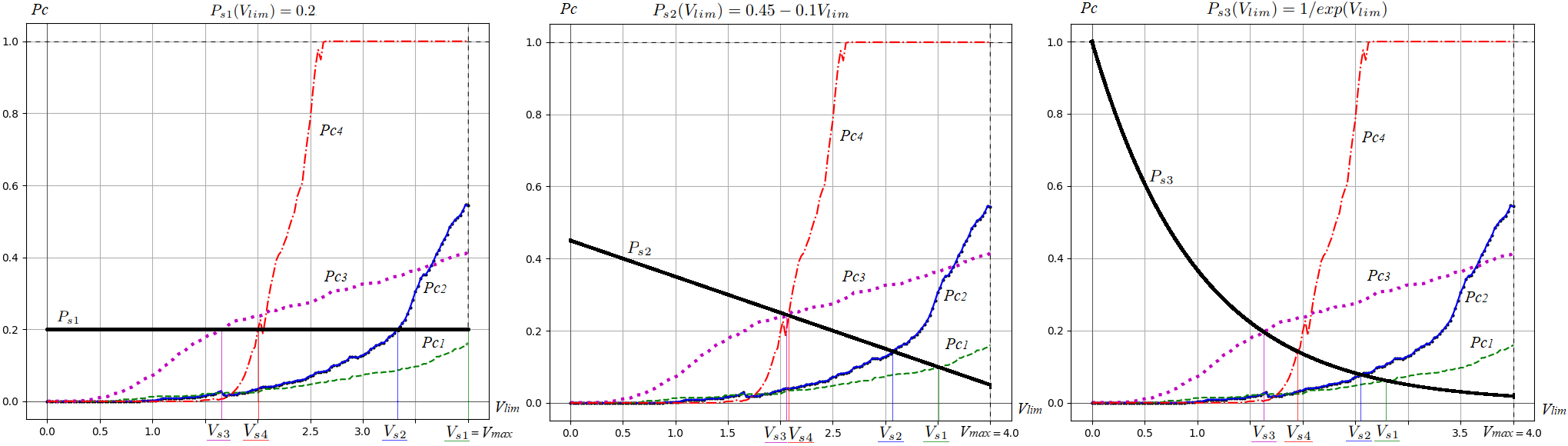}
    \caption{Types of collision probability threshold function and corresponding maximum safe speed values for few $ P_C(V_{\mathrm{lim}}) $ charts (constant function on the left, decreasing linear function in the center, decreasing exponential function on the right). }
    \label{fig:chart_with_threshold}
\end{figure*}

Fig.~\ref{fig:result_main} in the center shows the dependence of collision probability $P_C$ on speed limit $ V_{\mathrm{lim}} $ over time. Data are recorded from real passage. Along the X-axis the route passage time is shown. The Y-axis represents the sequence of speed limit values from 0 to the maximum possible on this route -- $4$ m/s. The color (according to the scale on the left) shows the probability of a collision within the prediction horizon $ \tau $ for the entire range of speed limits. Since deviations from the monotonic increase in the $ P_{C} (V_{\mathrm{lim}}) $ function are insignificant, it is possible to use fast search algorithms on ordered data to determine the maximum safe speed $ V_s $. Also probability threshold function $ P_s $ must be monotonically decreasing or constant. It is noticeable that $ P_{C} (V_{\mathrm{lim}}) $ varies greatly over time. Function rapid changes occur mainly at the moments of localization resampling (resampling of particles in the Monte Carlo localization algorithm). To avoid unwanted jerks (alternating high accelerations and braking) of the vehicle, several probability threshold functions $ P_{si} $ were analyzed.

The lower part of Fig.~\ref{fig:result_main} shows 3 dependences of the collision probability $ P_C $ on the speed limit $ V_{\mathrm{lim}} $ over time from the same passage, where black indicates forbidden speed limits, the collision probability of which exceeds the threshold according to $ P_s $ function. 
For clarity, the same functions $ P_{si} $, as well as the corresponding maximum safe speeds $ V_{si} $ for typical $ P_{C} (V_{\mathrm{lim}}) $ dependencies are shown in Fig.~\ref{fig:chart_with_threshold}. Back to Fig.~\ref{fig:result_main}, it is noticeable that when using the decreasing functions $ P_{s2} $ or $ P_{s3} $, the maximum safe speed limits $ V_{si} $ change much smoother than when applying the constant threshold function $ P_{s1} $, motion becomes less jerky. Moreover, the decreasing nature of the collision probability threshold function $ P_s $ is justified by the fact that on the high speeds even small threshold probability cannot be neglected which means the threshold value should decrease with the increase of speed. 

In the central graph in Fig.~\ref{fig:result_main} after the 200-th second, a region is noticeable where, at low speeds, the collision probability is close to zero, but when the speed increases it abruptly becomes close to 1. The situations describing these cases are shown in Fig.~\ref{fig:result_main} in the upper right and upper center images, respectively. We can see that the low probability at low speeds is explained by the high accuracy of localization in this section of the circular motion. A high probability is caused by the fact that the trajectory control system is not capable of completing a maneuver of a turn at such speeds which leads to a collision. In other areas, the increased probability of a collision is due, for the most part, to high ego-pose uncertainty, as in the upper left image of Fig.~\ref{fig:result_main}.

\section{CONCLUSION}

In this paper we proposed the safe speed control system for autonomous vehicle, based on real-time calculation of safe speed limit. The main contribution is the method for collision probability estimation that takes into account the ego-pose uncertainty for calculation of collision probability with both static and dynamic obstacles. Experiments showed that the proposed system allows avoiding collisions by decreasing vehicle speed.

\addtolength{\textheight}{-1cm}

The proposed approximations for taking into account the uncertainty of future trajectories made it possible to obtain an upper bound of the collision probability as the worst-case scenario with a minimum number of computational operations. Also, the use of the discovered property about the monotonically increasing function of the dependence of the collision probability on the speed limit allows us to reduce the computational complexity by an order of magnitude.
The proposed method allows not only to prevent a collision with a specific obstacle but can also be used to control the quality of the localization system, i.e. with a high uncertainty of ego-pose, the number of predicted collisions with arbitrary obstacles increases, as a result of which vehicle will slowed down or even stopped until the pose estimation will become more accurate.

In the future, the proposed method of collision probability estimation can be expanded by taking into account the error model of dynamic obstacle detector. It is also possible to take into account prediction of trajectories of dynamic obstacles.

\bibliographystyle{IEEEtran}
\bibliography{mybib}

\begin{thebibliography}{10}
\providecommand{\url}[1]{#1}
\csname url@rmstyle\endcsname
\providecommand{\newblock}{\relax}
\providecommand{\bibinfo}[2]{#2}
\providecommand\BIBentrySTDinterwordspacing{\spaceskip=0pt\relax}
\providecommand\BIBentryALTinterwordstretchfactor{4}
\providecommand\BIBentryALTinterwordspacing{\spaceskip=\fontdimen2\font plus
\BIBentryALTinterwordstretchfactor\fontdimen3\font minus
  \fontdimen4\font\relax}
\providecommand\BIBforeignlanguage[2]{{%
\expandafter\ifx\csname l@#1\endcsname\relax
\typeout{** WARNING: IEEEtran.bst: No hyphenation pattern has been}%
\typeout{** loaded for the language `#1'. Using the pattern for}%
\typeout{** the default language instead.}%
\else
\language=\csname l@#1\endcsname
\fi
#2}}

\bibitem{harper2016cost}
C.~D. Harper, C.~T. Hendrickson, and C.~Samaras, ``Cost and benefit estimates
  of partially-automated vehicle collision avoidance technologies,''
  \emph{Accident Analysis \& Prevention}, vol.~95, pp. 104--115, 2016.

\bibitem{liu2018safe}
P.~Liu, R.~Yang, and Z.~Xu, ``How safe is safe enough for self-driving
  vehicles?'' \emph{Risk analysis}, 2018.

\bibitem{wang2019crash}
H.~Wang, Y.~Huang, A.~Khajepour, Y.~Zhang, Y.~Rasekhipour, and D.~Cao, ``Crash
  mitigation in motion planning for autonomous vehicles,'' \emph{IEEE
  Transactions on Intelligent Transportation Systems}, vol.~20, no.~9, pp.
  3313--3323, 2019.

\bibitem{lambert2008fast}
A.~Lambert, D.~Gruyer, and G.~Saint~Pierre, ``A fast monte carlo algorithm for
  collision probability estimation,'' in \emph{Control, Automation, Robotics
  and Vision, 2008. ICARCV 2008. 10th International Conference on}.\hskip 1em
  plus 0.5em minus 0.4em\relax IEEE, 2008, pp. 406--411.

\bibitem{hu2018dynamic}
X.~Hu, L.~Chen, B.~Tang, D.~Cao, and H.~He, ``Dynamic path planning for
  autonomous driving on various roads with avoidance of static and moving
  obstacles,'' \emph{Mechanical Systems and Signal Processing}, vol. 100, pp.
  482--500, 2018.

\bibitem{kuffner2000rrt}
J.~J. Kuffner~Jr and S.~M. LaValle, ``Rrt-connect: An efficient approach to
  single-query path planning,'' in \emph{ICRA}, vol.~2, 2000.

\bibitem{bopardikar2015multiobjective}
S.~D. Bopardikar, B.~Englot, and A.~Speranzon, ``Multiobjective path planning:
  Localization constraints and collision probability,'' \emph{IEEE Transactions
  on Robotics}, vol.~31, no.~3, pp. 562--577, 2015.

\bibitem{du2011probabilistic}
N.~E. Du~Toit and J.~W. Burdick, ``Probabilistic collision checking with chance
  constraints,'' \emph{IEEE Transactions on Robotics}, vol.~27, no.~4, pp.
  809--815, 2011.

\bibitem{janson2018monte}
L.~Janson, E.~Schmerling, and M.~Pavone, ``Monte carlo motion planning for
  robot trajectory optimization under uncertainty,'' in \emph{Robotics
  Research}.\hskip 1em plus 0.5em minus 0.4em\relax Springer, 2018, pp.
  343--361.

\bibitem{schmerling2016evaluating}
E.~Schmerling and M.~Pavone, ``Evaluating trajectory collision probability
  through adaptive importance sampling for safe motion planning,'' \emph{arXiv
  preprint arXiv:1609.05399}, 2016.

\bibitem{broadhurst2005monte}
A.~Broadhurst, S.~Baker, and T.~Kanade, ``Monte carlo road safety reasoning,''
  in \emph{IEEE Proceedings. Intelligent Vehicles Symposium, 2005.}\hskip 1em
  plus 0.5em minus 0.4em\relax IEEE, 2005, pp. 319--324.

\bibitem{liu2014incremental}
W.~Liu and M.~H. Ang, ``Incremental sampling-based algorithm for risk-aware
  planning under motion uncertainty,'' in \emph{Robotics and Automation (ICRA),
  2014 IEEE International Conference on}.\hskip 1em plus 0.5em minus
  0.4em\relax IEEE, 2014, pp. 2051--2058.

\bibitem{van2011lqg}
J.~Van Den~Berg, P.~Abbeel, and K.~Goldberg, ``Lqg-mp: Optimized path planning
  for robots with motion uncertainty and imperfect state information,''
  \emph{The International Journal of Robotics Research}, vol.~30, no.~7, pp.
  895--913, 2011.

\bibitem{vitus2011closed}
M.~P. Vitus and C.~J. Tomlin, ``Closed-loop belief space planning for linear,
  gaussian systems,'' in \emph{Robotics and Automation (ICRA), 2011 IEEE
  International Conference on}.\hskip 1em plus 0.5em minus 0.4em\relax IEEE,
  2011, pp. 2152--2159.

\bibitem{houenou2014risk}
A.~Hou{\'e}nou, P.~Bonnifait, and V.~Cherfaoui, ``Risk assessment for collision
  avoidance systems,'' in \emph{17th International IEEE Conference on
  Intelligent Transportation Systems (ITSC)}.\hskip 1em plus 0.5em minus
  0.4em\relax IEEE, 2014, pp. 386--391.

\bibitem{patil2012estimating}
S.~Patil, J.~Van Den~Berg, and R.~Alterovitz, ``Estimating probability of
  collision for safe motion planning under gaussian motion and sensing
  uncertainty,'' in \emph{Robotics and Automation (ICRA), 2012 IEEE
  International Conference on}.\hskip 1em plus 0.5em minus 0.4em\relax IEEE,
  2012, pp. 3238--3244.

\bibitem{schreier2016integrated}
M.~Schreier, V.~Willert, and J.~Adamy, ``An integrated approach to
  maneuver-based trajectory prediction and criticality assessment in arbitrary
  road environments,'' \emph{IEEE Transactions on Intelligent Transportation
  Systems}, vol.~17, no.~10, pp. 2751--2766, 2016.

\bibitem{mehta2018c}
D.~Mehta, G.~Ferrer, and E.~Olson, ``C-mpdm: Continuously-parameterized
  risk-aware mpdm by quickly discovering contextual policies,'' in \emph{2018
  IEEE/RSJ International Conference on Intelligent Robots and Systems
  (IROS)}.\hskip 1em plus 0.5em minus 0.4em\relax IEEE, 2018, pp. 7547--7554.

\bibitem{annell2016probabilistic}
S.~Annell, A.~Gratner, and L.~Svensson, ``Probabilistic collision estimation
  system for autonomous vehicles,'' in \emph{2016 IEEE 19th International
  Conference on Intelligent Transportation Systems (ITSC)}.\hskip 1em plus
  0.5em minus 0.4em\relax IEEE, 2016, pp. 473--478.

\bibitem{lienke2018ad}
C.~Lienke, M.~Keller, K.-H. Glander, and T.~Bertram, ``An ad-hoc sampling-based
  planner for on-road automated driving,'' in \emph{2018 21st International
  Conference on Intelligent Transportation Systems (ITSC)}.\hskip 1em plus
  0.5em minus 0.4em\relax IEEE, 2018, pp. 2371--2376.

\end{thebibliography}

\newpage

\copyright{} 2020 IEEE.  Personal use of this material is permitted.  Permission from IEEE must be obtained for all other uses, in any current or future media, including reprinting/republishing this material for advertising or promotional purposes, creating new collective works, for resale or redistribution to servers or lists, or reuse of any copyrighted component of this work in other works.

\end{document}